# Towards eXplicitly eXplainable Artificial Intelligence

V. L. Kalmykov[1,*], L.V. Kalmykov[2]

[1] Institute of Cell Biophysics of the Russian Academy of Sciences, 3 Institutskaya street, Pushchino, Moscow region, 142290, Russia

[2] Institute of Theoretical and Experimental Biophysics of the Russian Academy of Sciences, 3 Institutskaya street, Pushchino, Moscow region, 142290, Russia

[*] Corresponding author.
E-mail address: vyacheslav.l.kalmykov@gmail.com (V.L. Kalmykov).

**Graphical abstract**

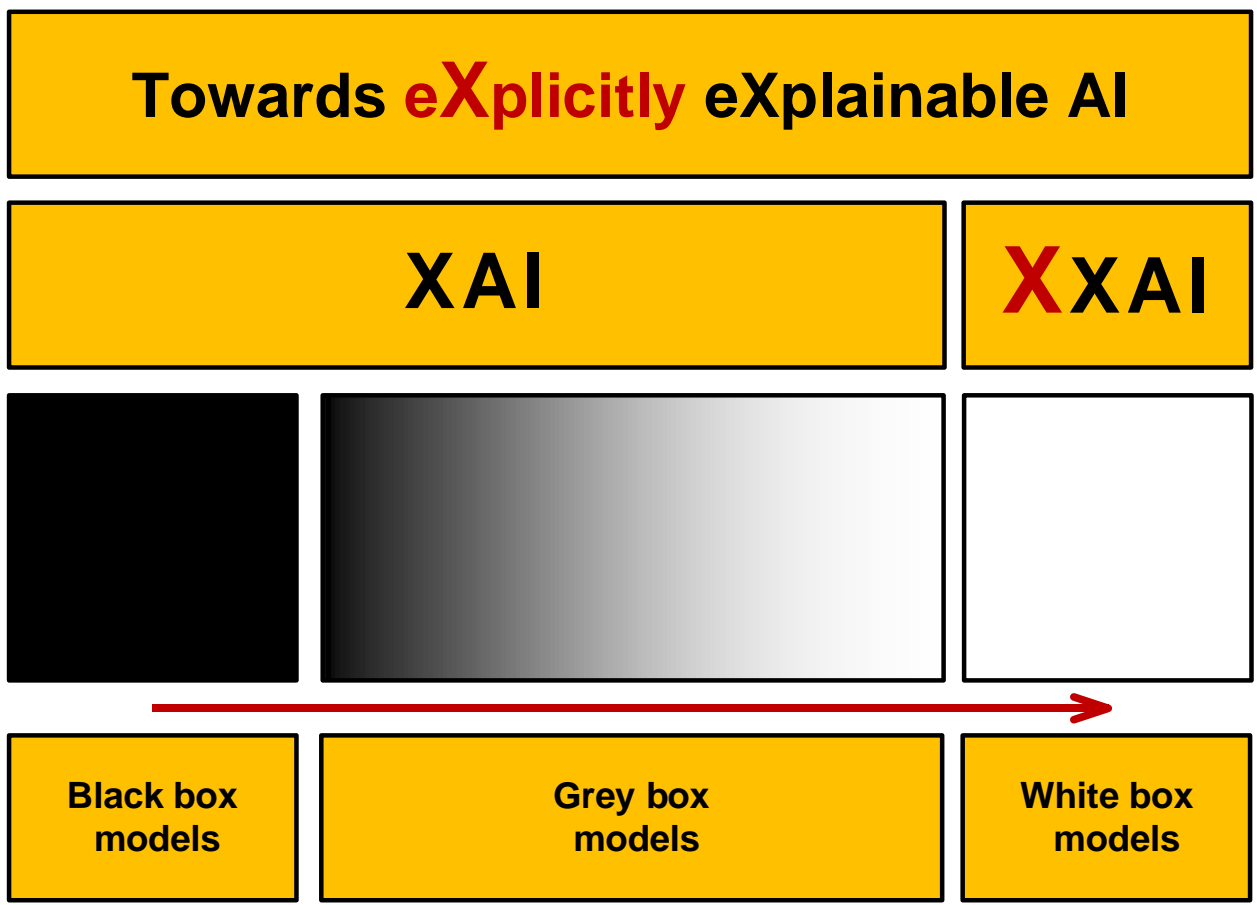

# HIGHLIGHTS

- eXplicitly eXplainable AI (XXAI) solves the AI black box problem
- Deterministic Logical Cellular Automaton implements general-purpose symbolic AI
- Transparent white box AI ensures the trustworthiness of automated decisions
- Cellular automaton fuses local causal inferences into a one global decision
- Four barriers to widespread adoption of symbolic AI have been overcome

**Abstract**

There are concerns about the reliability and safety of artificial intelligence (AI) based on sub-symbolic neural networks because its decisions cannot be explained explicitly. This is the black box problem of modern AI. At the same time, symbolic AI has the nature of a white box and is able to ensure the reliability and safety of its decisions. However, several problems prevent the widespread use of symbolic AI: the opacity of mathematical models and natural language terms, the lack of a unified ontology, and the combinatorial explosion of search capabilities. To solve the black-box problem of AI, we propose eXplicitly eXplainable AI (XXAI) - a fully transparent white-box AI based on deterministic logical cellular automata whose rules are derived from the first principles of the general theory of the relevant domain. In this case, the general theory of the domain plays the role of a knowledge base for deriving the inferences of the cellular automata. A cellular automaton implements parallel multi-level logical inference at all levels of organization - from local interactions of the element base to the system as a whole. Our verification of several ecological hypotheses sets a precedent for the successful implementation of the proposed solution. XXAI is able to automatically verify the reliability, security and ethics of sub-symbolic neural network solutions in both the final and training phases. In this article, we present precedents for the successful implementation of XXAI, the theoretical and methodological foundations for its further development, and discuss prospects for the future.

# 1. Introduction

### 1.1. Concerns about reliability and security of contemporary artificial intelligence

Machine learning systems based on artificial neural networks have achieved a number of breakthroughs by combining huge machine learning data sets with the enormous processing power of modern computers. However, it is well known that artificial intelligence (AI) based on neural networks has a "black box" nature and its decisions cannot be explicitly explained. Concerns about the trustworthiness and security of their solutions have led to the AI black box problem. Mathematical models of complex systems that underlie the most successful AI systems are based on the methods of continuous mathematics and probability theory. These models are of the "black box" type. At the same time, our world clearly consists of discrete systems. The computers we use are also discrete in nature. Much effort has gone into making discrete computers suitable for continuous mathematical modeling. Opaque mathematical models of the black box type are not able to take into account the local conditions of the behavior of the subsystems of the objects under study and are based rather on fitting the model parameters to the observed phenomenology or known facts. The conclusions of opaque models about the internal mechanisms of complex systems have to be made as indirect subjective interpretations, which gives rise to paradoxes and complicates the understanding of the phenomena under study. All this is true for the most successful neural network AI of the GPT type today, based on the Transformer model. These new approaches to deep learning have overshadowed all earlier approaches to AI (Korngiebel and Mooney 2021). However, Transformer models, including GPT versions, are opaque and do not yet create human-interpretable semantic representations that underlie generated sentences and documents. IN Transformer models are not are used clean rational brain teaser operations such like AND, OR, XOR and they do not rely on production rules such as "if-then" rules. The algorithms of the Transformer model allow it to make decisions about which word or phrase should follow the previous words or phrases in order to generate the text with the highest likelihood of believability. At the same time, AI based on the Transformer architecture has a higher degree of explainability of the decisions made compared to classical neural networks. This is because Transformer models use an attention mechanism that allows you to understand what input was used to generate each part of the output. Thus, Transformer models can provide some explanations and interpretations of their predictions. However, as with other deep learning models, a white-box transparency is still an elusive goal, and some aspects of the model may not be explicitly explainable. Decisions of machine learning systems are made on an irrational, sub-symbolic basis. We cannot know the logic of the decisions they make, and we learn these decisions only after the fact from the behavior of the AI-controlled system. This lack of transparency in decision making can lead to errors, especially in unfamiliar, unexpected, and unusual situations. Douglas Lenat , who has led the development of the largest knowledge base of common sense "CYC" since 1984, compares modern AI with a patient whose left hemisphere of the brain, which is responsible for logical thinking, is completely damaged (Lenat 2019). These circumstances are unacceptable for a number of sensitive industries – medicine (Gorban et al. 2021; Rudin 2019), law, transport, insurance, security systems, including military ones. In 2016, DARPA, the U.S. Department of Defense's Defense Advanced Research Projects Agency, initiated the Explainable Artificial Intelligence (XAI) project (Gunning and Aha 2019). The purpose of the XAI Project is to "open the black box" of machine learning in order to explain the reasons for the decisions made by AI, to increase the security and reliability of these decisions, and to increase the trust of the end user in them. The strategic goal of the project is to create autonomous intelligent systems that perceive, learn, make decisions and act independently, symbiotically combining machine learning with other artificial intelligence technologies. In subsequent years, a lot of research on "XAI" appeared, but so far, they are not devoted to opening the black box of neural network AI, but to assurances that black box systems can be trusted (Adadi and Berrada 2018; Barredo Arrieta et al. 2020; Gorban et al. 2021; Arras et al. 2022; Borrego-Díaz and Galán-Páez 2022; Buijsman 2022). These are attempts to legitimize black box AI, not to make AI truly transparent. The black box of neural network AI remains closed. Rather, this approach to the AI black box problem seeks to perpetuate bad practice and has the potential to cause great harm to society (Rudin 2019). Dissatisfaction with the results of attempts to crack the AI black box in the XAI project led to the need to go beyond the scope of that project and move directly to ensuring transparency, efficiency, and reliability - (Holzinger et al. 2022).

### 1.2. From trying to crack the black box of contemporary AI to creating its white box partner - an explicitly explainable AI (XXAI)

Attempts to crack the black box of contemporary AI and find a logical explanation for its decision-making processes are reminiscent of attempts to capture thoughts in the human brain with a scalpel and electrodes. We believe that instead of trying to crack the black box of AI, we need to create its fully transparent partner, which is inherently a white box. Really explicable and transparent artificial intelligence with available decision-making algorithms should use white-box mathematical models (Kalmykov and Kalmykov 2021; Kalmykov and Kalmykov 2024). Unlike black-box AI, white-box AI decision algorithms are human-understandable and can be based on ethical, usefulness, and safety principles. In the future, fully transparent AI based on logical cause-and-effect reasoning can effectively complement non-transparent machine learning to form a hybrid neuro-symbolic AI (Goertzel 2012; Calegari et al. 2020; Ilkou and Koutraki 2020; Ebrahimi et al. 2021; Díaz-Rodríguez et al. 2022; Stehr et al. 2022; Md Kamruzzaman et al. 2022; Tarek et al. 2022; Hochreiter 2022; Kautz 2022; N'unez-Molina et al. 2023).

Today, the learning process of deep neural network transformers using the Reinforcement Learning from Human Feedback method is done through human mediation (Santacroce et al. 2023). or using the Reinforcement Learning with AI Feedback (RLAIF) method mediated by another transformer neural network (Lee et al. 2023). Human control of the neural network training process is expensive and time-consuming. In addition, human control over AI decisions makes the very idea of automation largely meaningless, and human capabilities in terms of speed of decision-making and volume of information processed are significantly inferior to the capabilities of modern computers. We believe that in the future of hybrid artificial intelligence, symbolic AI will take control of the learning and decisions of sub-symbolic AI. XXAI can automatically verify the reliability, safety, and ethicality of sub-symbolic neural network AI decisions during both the final and training phases.

Norbert Wiener, in his Cybernetics, used the terms "black box" and "white box" to characterize types of mathematical models of complex systems (Wiener 1948). Although black-box mathematical models are now well known and successfully implemented, the implementation of white-box models, especially in AI, still raises questions. We treat phenomenological models of complex systems as black boxes and mechanistic models of complex systems as white boxes (Kalmykov and Kalmykov 2015c, b). Fully mechanistic models of complex systems are completely discrete and deterministic. They allow us to directly gain a mechanistic insight into a system under study (Kalmykov and Kalmykov 2013; Kalmykov and Kalmykov 2016; Kalmykov and Kalmykov 2021).

### 1.3. On symbolic artificial intelligence

Symbolic AI is a completely transparent white box. It allows logical verification of the security and reliability of its solutions. Symbolic AI works logically with symbols. Each symbol in symbolic AI represents a specific concept. This provides a causal understanding of its decisions. Understanding is our ability to logically reproduce the decision process in our imagination, in text, or in a computer. Such reproduction begins with a specific element base of the model and takes into account all local connections of all elements within the model, allowing the entire model to be reproduced as a whole. Decision clarity is characterized by its transparency, which allows symbolic AI to be viewed as a white box. At the moment, the most successful implementations of artificial intelligence are its sub-symbolic forms based on neural networks. Sub-symbolic AI is a black box because humans cannot logically trace the cause-and-effect mechanisms of their decisions. This is the black box problem of modern AI (Castelvecchi 2016). To be as good at solving a variety of complex problems as humans are, machines must learn to build logical cause-and-effect models of the environment and navigate different contexts. Building logical cause-and-effect models is a specialization of symbolic AI.

In the early days of AI, logical and symbolic reasoning methods predominated (Swartout et al. 1991; Lent et al. 2004). AI of a symbolic-logical nature is known as Good Old-Fashioned Artificial Intelligence (GOFAI) (Boden, 2014; Haugeland, 1985). The birth of artificial intelligence and expert systems in the 1960s and 1970s led to significant advances in symbolic AI. From the 1970s to the 1990s, these methods encountered a number of problems that were never overcome. To this day, the widespread use of symbolic AI is hampered by four fundamental barriers that were identified half a century ago:

1. *Operational opacity of the mathematical apparatus*. This problem arose due to the lack of transparency of most mathematical methods, which are fundamentally unable to control the local interactions of local interactions of subsystems of the modeled system (Kalmykov and Kalmykov 2015c; Kalmykov and Kalmykov 2016; Kalmykov and Kalmykov 2015b, 2021).

2. *Semantic (ontological) opacity* (implicitness, non-strictness and polysemousness) *of natural language terms* that lead to *general abstract ontology* errors in automatic inference (Alibali and Koedinger 1999; Brinck 1999; Dienes and Perner 1999; Smith 2001; Dreyfus 1972; Dreyfus et al. 1986; Fjelland 2020).

3. *Lack of a general abstract ontology of our world.* This does not allow realizing a common scientific understanding, agreeing on the meanings of concepts, and formulating universal ethical principles that are common to humans and AI. This general top-level ontological knowledge should be formulated in explicit, easily applicable definitions (Dreyfus 1972; Nickel 2010; Díez et al. 2013).

4. *Irresistible combinatorial explosion.* Often an irresistible combinatorial explosion that occurs when sorting through logically possible options in search of the best solution (Lighthill 1973). Such an enumeration may require more time than the time of the heat death of our universe (Lenat 2019).

Failure to solve any of these problems will negate the widespread use of symbolic AI. The safe, secure and trustworthy development and use of artificial intelligence is receiving great attention (Biden 2023; The Bletchley Declaration by Countries Attending the AI Safety Summit, 1-2 November 2023). The quest to ensure the reliability and security of the most effective AI faces the challenge of ensuring the algorithmic transparency of the decisions made. Only symbolic AI, through its full algorithmic transparency, can implement automatic protection of AI decisions to keep pace with existing and emerging threats. Let us characterize these obstacles and ways to overcome them in more detail.

## 2. Main hypothesis

Explicitly explainable AI (XXAI) is possible as a white box solution to the AI black box problem that overcomes all four barriers listed here on the basis of deterministic logical cellular automata whose rules are derived from the axioms of the general physical theory of the relevant domain:

1. *Operational (syntactic) transparency* of the mathematical apparatus is achieved by cellular-automatic cause-and-effect fusion of all local cause-and-effect inferences into a global causal conclusion. These global conclusions are accepted at each iteration of the cellular automaton.
2. *Semantic (ontological) transparency* is achieved through the use of explicit concepts both at the level of local decisions about the behavior of individuals of the element base, and at higher levels of organization, including the behavior of the system as a whole through the use of the axioms of the general theory of the relevant domain.
3. *The general abstract ontology of our world* is based on general axiomatic theory of life (Kalmykov 1997; Kalmykov 2012).
4. *The insurmountable combinatorial explosion in the search for options i*s eliminated by direct cause-and-effect cellular automata aggregation of local solutions into a single global solution. The aggregation is based on cellular automata neighborhood and cause-and-effect rules of state transitions. Scenarios for the implementation of these cause-and-effect rules are developed on the basis of the axioms of the general physical theory of the relevant domain.

XXAI in a broader sense can be a neuro-symbolic AI, where the training and decisions of sub-symbolic AI with statistical machine learning are controlled by fully transparent rational symbolic intelligence.

## 3. Overcoming the barriers to widespread use of symbolic AI

To create a general-purpose symbolic AI, it is necessary to implement a transparent mathematical algorithm, i.e. the ability of the AI to make decisions algorithmically, hierarchically combining individual logical steps into a common output. The ability to implement a complex algorithm as a whole, starting from its elementary components, is based on the knowledge of the mechanism of operation of the system. This knowledge is the foundation of our understanding. It is this understanding of AI decision algorithms that underlies their explicability. Next, we'll take a closer look at ways to overcome the barriers to widespread use of symbolic AI that can work with knowledge and make decisions that we can fully understand.

### 3.1. Overcoming operational opacity of basic mathematical models

The paper (Setzu et al. 2021) points out the need to create global frameworks that integrate local cause-and-effect interactions to solve the black box problem in AI. Cellular automata allow you to create a transparent model of the system, providing integration of the behavior of the system elements through their local interactions. *Logical deterministic cellular automata are an uncontested basic tool for ensuring the operational transparency of mathematical models for AI.* The transparency of such models is ensured by the ability to control the states and relationships of basic elements within the modeled system through a local cellular automaton neighborhood. We can talk about a special cellular-automaton method of symbolic thinking, which combines the following two points: (1) space modeling by a grid of discrete states of the element base of the system; (2) modeling time by successive states of system elements according to the rules of local dependencies specified by the general theory of the corresponding subject area.

Fig. 1 shows our classification of mathematical models according to the degree of their transparency, depending on the type of mathematical methods used (Kalmykov and Kalmykov 2013, 2015c).

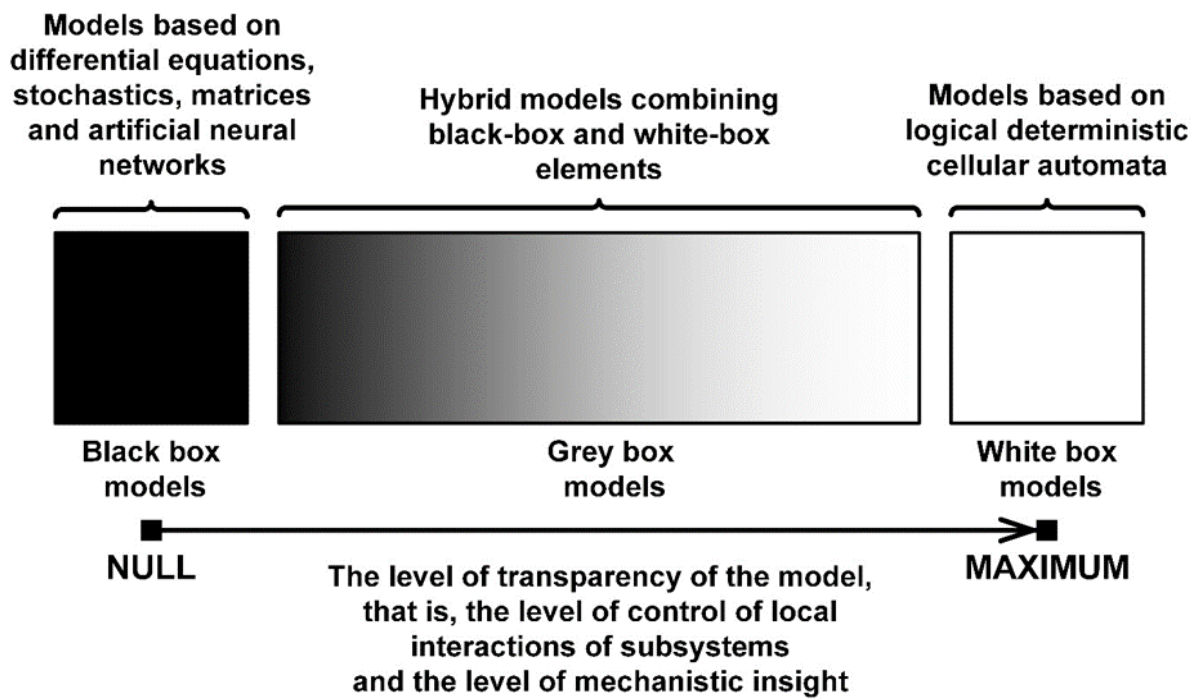

**Fig. 1. Universal classification of mathematical models of complex systems according to the degree of their transparency depending on the type of used mathematical methods.** This is a schematic representation of black box, gray box and white box type models with indication of the level of control over interactions of subsystems of the complex system.

It follows from the hypothesis presented by us that a completely transparent model of a system may be created by logical deterministic cellular automata, the rules of which are based on the general physical theory of the subject area of the system being modeled.

### 3.2. Overcoming semantic opacity (implicitness, non-*strictness and polysemy) of natural language terms*

Natural language concepts present inherent limitations, including their implicit nature, insufficient semantic coordination, and potential errors and contradictions in conventional ontologies, which impede the development of reliable automated reasoning systems.

*The semantic explicitness of terms can be achieved only within the framework of the general theory of the relevant subject area. Within the framework of such a theory, the terms not only have a strictly defined meaning, but their meanings are fully consistent with each other at all levels of the organization of the model.*

To ensure correct automatic inference, it is necessary to carry out the procedure of preliminary theoretical purification (idealization) of the used concepts of natural language, applying the principles of classical scientific rationality. The mysterious power of mathematics (Wigner 1960) lies precisely in the fact that they are based on a strict axiomatic definition of basic concepts, rules for operating these basic concepts, and on careful coordination of these concepts into a single integral theory. Such a theory is the most perfect knowledge base for recreating the objects of the relevant subject area in the minds of students, in computer models and in technical objects, including at a new, more advanced level.

To understand means to reproduce an object in the imagination or in a computer, starting from its elemental base. Scientific rational knowledge provides such an opportunity. Here a cultural problem arises - rational knowledge has a supersensory abstract nature, and today the acquisition of experimental data and their statistical processing are at the forefront of science. Abstract results are often considered speculative and are more likely to be disapproved of by reviewers. The editor-in-chief of the journal Nature, John Maddox, noted that the publication of the famous article by Watson and Crick on the structure of DNA today would be impossible because. reviewers would definitely consider it speculative (Maddox 1989). This article was published without peer review as self-evident (Maddox 2003).

### 3.3. Creation of a common abstract ontology of our world

We are convinced that the creation of an ontology of transparent AI is possible only on the basis of a general abstract theory of the relevant subject area. For successful work in this direction, the culture of creating such general abstract theories must be comprehensively developed. The experience of mathematics and programming in formal languages has shown that logical operations with symbols corresponding to completely explicit concepts give semantically correct results. To turn partially implicit concepts into explicit ones, it is necessary to build a general axiomatic theory of the subject area of the modeled object. Such a detailed specification of conceptualization is the construction of an ontology of a given subject area (Gruber 1995). Each specific axiomatic theory makes it possible to build an adequate logical model in its subject area. For general artificial intelligence, it will be necessary to create a general ontological theory that is equally applicable to any subject area. In our opinion this problem coincides with the general formulation of Hilbert's sixth problem on the existence of mathematical foundations of the physical world. At present, there is no formal theory in which to express the analysis of the means and ends used both in ordinary life and in physics. The creation of such a formal theory of the most general nature is one of the key promising tasks for ensuring the reliability of symbolic AI (McCarthy 1968). A common abstract ontology provides a unified conceptual framework for different domains. For our research, the general semantic basis was the abstract theory of living systems (Kalmykov 1997; Kalmykov 2012). Based on this theory, we formulated a general axiomatic theory of the ecosystem for the study of interspecific competition, which is presented in our previous works (Kalmykov and Kalmykov 2013, 2015b; Kalmykov et al. 2017) and in greater detail in our articles (Kalmykov and Kalmykov 2021,2024).

### 3.4. Solving the combinatorial explosion problem

The fourth problem faced by the developers of symbolic intelligence in the 70s was the often-insurmountable combinatorial explosion that occurs when a complete enumeration of possible options in search of the best solution (Lighthill 1973). To overcome the problems of complete enumeration of the space of possible solutions, we use the direct derivation of the optimal solution, starting from the local rules for the optimal behavior of micro-objects of the simulated system. Logical local rules for the behavior of micro-objects are given by the cellular automaton neighborhood and glue the behavior of micro-objects into the behavior of an integral system. The evolution time of such a model as a whole is directly proportional to the size of the system and the length of the rules of behavior used. Local cellular automaton rules should be based on the first principles of the general theory of the subject area. Then the operational cellular automaton model building from the bottom up ("bottom–up") is complemented by the simultaneous semantic model building from the top down ("top-down"). As a result of combining the possibilities of the general abstract theory of the subject area of the modeled object with the possibilities of productive cellular-automatic inference, automatic deductive inference is carried out. A detailed description of the method is presented in our work (Kalmykov and Kalmykov 2021).

## 4. Discussion of the problems of symbolic AI and ways to overcome them

### 4.1. Features and capabilities of cellular automata

"Cellular automata are mathematical idealizations of physical systems in which space and time are discrete, and physical quantities take on a finite set of discrete values" (Wolfram 1983). The system is modeled by a cellular automaton as an iteratively changing structured set of homogeneous elements. Each element of this set can be in at least two states. The states of all elements of this set change step by step and simultaneously according to certain logical rules. The rules determine what state each given element will go to at the next step and depend on what state the element itself is in at the moment and what states its neighbors are in, given by the local cellular automaton neighborhood. Homogeneous elements, coordinated with each other locally throughout the space of a structured set, are always and uniquely glued together into an integral global system. The global behavior of the system is generated from the local rules for changing the states of its micro-objects.

Formally, a cellular automaton can be defined as a set of five objects:

    1. Regular homogeneous lattice of nodes (each site models a micro-object);

    2. A finite set of possible node states;

    3. Initial pattern of states of lattice nodes.

4. A neighborhood consisting of a node and certain neighboring nodes. This set of nodes affects each node transition to the next state;

5. The function of transition to the next state - the rules of iterative change of states of each node without exception, which are applied to all nodes of the lattice simultaneously. These rules take into account the state of the neighborhood of each node. All nodes have the same update rules. The node state iterations model the timeline. The logical causal structure that governs the transition between node states is based on if-then rules.

Each node of a cellular automaton is a finite automaton, and the cellular automaton as a whole is a polyautomaton that co-organizes the behavior of the finite automata located in the nodes of its lattice. The cellular automaton implements automatic inference as hyperlogic, since the cellular automaton logic is executed simultaneously for all nodes at each iteration. As a result, the cellular automaton implements knowledge-based inference. The behavior of the micro-objects of the simulated system is logically glued together into a single integral macro-object with the help of local interactions mediated by the cellular automaton neighborhood. It is this kind of modeling that logically reproduces the integration of subsystems into an integral system that allows us to speak about the white-box transparency of the models.

### 4.2. Is our Universe a cellular automaton?

*The deterministic dynamics of cellular automata* appears to be a candidate for the fundamental laws of physics, including quantum theory ('t Hooft 2020). The effectiveness of cellular automaton modeling can be related to the fact that the fundamental laws of nature are of a cellular automaton nature, and the Universe itself is a cellular automaton. Such a hypothesis was put forward earlier by Konrad Zuse in his book "Computing Space" (Zuse 2012). Zuse calls for an "automatic way of thinking" in the field of physics, which implies that a physical model is considered in terms of a sequence of states following each other according to predetermined rules. Following Zuse, Edward Fredkin believed that the Universe is a computer program that develops on a grid in accordance with a simple rule (Fredkin 1990). If we could only define the right grid type and the right rule, Fredkin says, we could model the universe and all of physics. Fredkin insists that the rule governing the universe's cellular automaton should be simple—just a few lines of clean and elegant code. Continuing the ideas of K. Zuse and Fredkin, Stephen Wolfram made attempts to find a digital cellular automaton code of the Universe . Eighteen years ago, he tried to find this code by enumerating the rules and patterns of states of lattice nodes of simple cellular automata (Wolfram 2002). In recent years, Wolfram has supplemented these searches by enumerating the rules of transformation graphs of sets of relations of identical but labeled discrete elements (Wolfram 2020; Wolfram and Wolfram Research (Firm) 2020). The code of the Universe has not yet been found, and Wolfram's project to search for a fundamental theory of physics has been criticized (Becker 2020). It is unlikely to find the rules for the creation of the universe just by combining numbers. Even if cellular automata and graphs are used. We share the idea that our world is based on cellular automaton logic, but we believe that a more meaningful search for an appropriate fundamental theory is needed. It is necessary to find the first principles (objects and axioms) of the general physical theory and put these principles into the rules of the cellular automaton of our Universe. Without creating a common ontology of our world, it is impossible to invest in AI algorithms for expedient, safe and useful behavior.

### 4.3. Cellular-automaton nature of board games on a cellular field

The use of the cellular automaton approach for modeling has been implemented in ancient board games. These games develop logical and imaginative thinking skills in the field of strategy and tactics. Chess is the most famous logic game that simulates the battle of two armies on the field. The most ancient of the currently known strategic games on the cellular field are senet and the royal game of Ur. The game of senet was invented in ancient Egypt about 5500 years ago. The royal game of Ur was invented in Sumer at least 4400 years ago.

We believe that in order to create a board game on a cellular field, it is first necessary to create a theory based on the first principles. To create a theory means to define the basic objects of the theory (pieces, board, starting pattern) and the axioms of the behavior of these basic objects (rules of the game). This classic rational approach to modeling emerged spontaneously and independently in different civilizations. The prevalence of logical board games on the cellular field in all advanced civilizations testifies to their unique effectiveness for modeling complex systems and the exciting interest that they aroused among the players, providing full visibility and transparency of all events on the field. We consider these games to be cellular automata because they have similar features: (1) the field consists of ordered cells, (2) each cell can be in a finite number of known states, (3) the pieces have specific local neighborhoods, (3) there is an initial field pattern (4) transitions between cell states are implemented iteratively according to certain logical rules. Given the prevalence of cellular automaton stereotypes in board games of different civilizations, it can be assumed that cellular automaton thinking is natural for humans. Perhaps our thinking uses cellular automaton modeling of the world we live in. At the same time, human imagination is not enough to use all the game pieces of the opposing sides at each iteration of the game. As a result, the game is played by alternating single moves of individual pieces. Computers opened up the possibility of modeling changes in the states of all cells of the field at each iteration, which increased the realism of the models, since in nature, many processes occur simultaneously. The board games on a cellular field create a simulation of reality. Their spontaneous cellular automata nature potentially confirms the idea that the Universe is a cellular automaton.

### 4.4. From the metaphors of the Game of Life to precedents for the implementation of completely transparent models

The computer cellular automaton game "Life" (Game of Life) by John Conway (Gardner 1970). The player arbitrarily creates a pattern of "live" (shaded) and "dead" (unshaded) cells on a two-dimensional field, and then only observes the evolution of the patterns. If the number of "living" neighboring cells is 2 or 3 out of 8, then at the next iteration the cell "survives", if not, then it dies. A "dead" cell "comes to life" at the next iteration if only it is surrounded by three "live" cells. The player hopes to get interesting dynamics during the evolution of the machine. Of course, this is not a model of a specific object of nature, but a very distant metaphor that can be attributed to abstract art. But this metaphorical model has analogies in many disciplines and allows you to reproduce dynamic processes that were not amenable to mathematical modeling before. For us, this model is important because, having taken its program (Naumov and Shalyto 2003) as a basis, we have replaced its rules taking into account the objects and axioms of the general theory of an ecosystem with competing species. The ecological model of resource competition that we have implemented has a completely transparent nature, unlike all available solutions, and has made it possible to solve a number of problems of theoretical ecology (Kalmykov and Kalmykov 2013, 2015b, 2021):

1. For the first time, completely transparent individual-oriented mechanisms for the formation of classical population growth curves, including the double S-shaped curve (Kalmykov and Kalmykov 2015c).

2. For the first time, a discrete model of population catastrophes has been created; it has been demonstrated that with an increase in the recovery time of ecosystem resources, a catastrophic death of the population occurs (Kalmykov and Kalmykov 2015a).

3. Mechanisms have been found for the indefinitely long coexistence of complete resource competitors under conditions under which coexistence was previously considered impossible. This solved the paradox of biodiversity and opened up new ways to preserve biodiversity (Kalmykov and Kalmykov 2013; Kalmykov and Kalmykov 2016; Kalmykov and Kalmykov 2015b).

4. Two contradictory hypotheses of limiting similarity and limiting difference of competing species were tested, the hypothesis of limiting similarity was rejected (Kalmykov and Kalmykov 2021).

5. For the first time, the competitive exclusion principle was rigorously tested and reformulated, and a generalized formulation of the competitive exclusion principle for an arbitrary number of resource competitors was given (Kalmykov and Kalmykov 2013, 2015b; Kalmykov and Kalmykov 2016).

For the first time, the general principle of competitive coexistence was formulated for an arbitrary number of resource competitors (Kalmykov and Kalmykov 2015b; Kalmykov and Kalmykov 2016).

### 4.5. The rules of our cellular automata models in brief

One individual can occupy only one microhabitat. The life cycle of an individual, like the state of regeneration of a microhabitat, lasts one iteration of the cellular automaton. All states of all sites have the same duration. Each individual of all species consumes the same number of identical resources, ie they are identical consumers. Such species are complete competitors. Individuals are immobile at the lattice sites and population waves propagate due to the reproduction of individuals. The neighborhood consists of a site and its given neighboring sites. All sites have the same update rules. The neighborhood models the meso-habitat of the individual and determines the number of possible descendants (fertility) of the individual. A more detailed description of our cellular automaton models can be found in our papers, especially those cited in this paper, especially in (Kalmykov and Kalmykov 2021), which contains a presentation of the general theory of an ecosystem with competing species and a basic source code.

### 4.6. Environmental aspect of symbolic AI decision modeling

The key feature of the proposed approach is that the problem situation solved by AI is modeled as a multi-agent ecosystem, the subsystems of which are isolated micro-ecosystems of individual agents. This isolation is actualized at the moment of making a decision about changing the state by each individual agent at each cycle of the cellular automaton. The meaning of this isolation is to take into account *all* the circumstances that are relevant for the expedient behavior of the agent at the moment of making this decision. The circumstances that are taken into account are part of the agent's cellular automaton neighborhood. For the future, we would like to note that not only each agent of the cellular automaton model can have a specific local neighborhood, but this neighborhood can also change depending on the circumstances that are identified by the sensory capabilities of the agent.

### 4.7. Cellular automata implement multilevel "top-down" and "bottom-up" approaches to modeling complex systems

Cellular automata are "bottom-up" models that generate global behavior from local rules (Silvertown et al. 1992). Through bottom-up inference, we were able to ensure that our model mechanistically constructs the ecosystem from the bottom up at each iteration, from the local conditions of each individual microhabitat and each individual species to the ecosystem as a whole. The rules of cellular automata are implemented at three levels of ecosystem organization. It is this multilevel modeling that makes it possible to speak about the white-box transparency of this method. The microlevel is modeled by a lattice site (microecosystem). The mesolevel of local interactions of a microecosystem is modeled by the neighborhood of cellular automata (mesoecosystem). The macro level (the entire ecosystem) is modeled by the entire lattice. The logical rules of the model include both "part-whole" relationships and "cause-effect" relationships. The whole is the entire cellular automaton (lattice shape and size, boundary conditions, cellular automaton neighborhood, general cellular automaton rules for state transitions), and the parts are the cells and their and their neighbors. This approach combines quantitative discreteness of microecosystems (environment) and their states with explicit consideration of their cause-effect relationships in modeling ecosystem dynamics.

Creation of a transparent mathematical model assumes knowledge of the algorithms for reproducing the system by logically combining its elements into an integral system. Transparency also implies knowledge of the algorithms that determine the mechanisms of the system's functioning. Understanding the mechanism implies the ability to create in the imagination or on a computer its transparent mathematical model, i.e., the transparent model of the system is its integral mechanism.

The key feature of the white-box transparency of the mathematical model is the requirement for the simultaneous implementation of cause-and-effect algorithms of the "part-whole" relationship both relatively "top-down" and relatively "bottom-up". This requirement applies equally to operational transparency and semantic transparency. If operational transparency is achieved using deterministic logical cellular automata, then semantic transparency must be provided by the general theory of the relevant subject area. The last requirement creates a problem because of the partial loss of the culture of rational thinking in modern positivist science. Simultaneous implementation of top-down and bottom-up approaches in multilevel cellular automata models eliminates the problem of confrontation between reductionist and holistic approaches.

## 5. Conclusions and prospects

### 5.1. Explicitly explainable AI (XXAI)

To overcome all four barriers to the widespread use of symbolic AI and to obtain a white box solution, we proposed the use of logically deterministic cellular automata whose rules are based on the general physical theory of the relevant domain. This is the main hypothesis of this paper. The comprehensive analysis of this hypothesis carried out here fully confirmed it and made it possible to make predictions regarding the prospects for implementing the proposed solution.

This approach opens up great prospects for various theoretical and applied areas. The artificial intelligence developers have gained a precedent for using fully transparent and fully explainable artificial intelligence to model the behavior of multi-particle and multi-level systems. Rational logical artificial intelligence brings with it the explainability, reliability, controllability and security. It is possible for AI not only to process data statistically, as in today's successful AI systems, but also to work directly with knowledge. This work provides the methodological foundation for a research and development program to create general-purpose symbolic AI as the explicitly explainable AI (XXAI).

### 5.2. Explicitly explainable AI (XXAI) Creation Program

We expect that the central result of this program will be the creation of a unified base of logical scenarios for rational agent behavior, based on first principles of general theories from various domains. Scientists and developers will be able to use these algorithms directly when creating symbolic AI systems. The main thing that will be needed to create such a foundation is a universal ontology of explicit knowledge (Dreyfus 1972; Nickel 2010; Díez et al. 2013). It is necessary to find a universal ethical foundation (goal function) that is common to all humans and all forms of AI. We propose to use the general theory of living systems as the basis for such a universal ontology (Kalmykov 1997; Kalmykov 2012). On the basis of axioms of general theories of different fields, a unified basis of first principles for fully transparent scenarios of rational behavior of any agent will be created. There is a certain difficulty in attracting the implementers of such a program due to the 200-year dominance of positivism in science. Rational knowledge is based on supersensible concepts. Supersensible concepts are insufficiently familiar objects for the work of modern scientists, who are accustomed to obtaining and publishing reliable experimental data. The path from experimental data to rational knowledge is through reflection, which is often perceived by reviewers as unacceptable speculation (Maddox 1989, 2003). The ship of modern science needs to make a course correction so that the traditions of classical principles of scientific rationality find their place in the way we know the world around us. Scientists should see themselves not only as miners extracting data, but also as Euclids creating new axiomatic theories of their fields. Of course, it is necessary to understand that axioms cannot be arbitrary, but only the result of a sincere, deep and honest immersion in the respective subject. Only when the assimilation of the facts of a given subject is sufficiently complete, it is possible to achieve its internal understanding through persistent reflection on the available facts and the explanatory concepts of its predecessors. This internal model of the domain is the basis for creating the corresponding basic objects and axioms of general theories. In addition to general theoretical work to ensure the explicitness of the concepts used, the program for the widespread use of symbolic intelligence should include the comprehensive development of cellular automaton modeling. Each simulated agent must be able to rationally correct its cellular automaton neighborhood before each iteration. It is necessary to ensure hierarchical network co-organization of agents by timely and rational assignment of neighborhoods of cellular automata by higher agents to lower ones. It is necessary to combine the algorithms of transparent AI with modern geographic information systems. The use of category theory, which has created a rich language for modeling complex systems, seems very promising. (Goguen 1991; Baez and Stay 2011; Tull et al. 2023).